\documentclass[letterpaper]{article} 
\usepackage{listings}
\lstdefinelanguage{pddl}{
  morekeywords={define,domain,problem,requirements,predicates,action,parameters,precondition,effect,init,goal,and,or,not,when,forall,exists,object,types,constants},
  sensitive=true,
  morecomment=[l]{;},
  morestring=[b]""
}
\usepackage{aaai25}  
\usepackage{times}  
\usepackage{helvet}  
\usepackage{courier}  
\usepackage[hyphens]{url}  
\usepackage{graphicx} 
\usepackage{subcaption}
\usepackage{float}
\usepackage{natbib}  
\usepackage{caption} 
\usepackage{algorithm}
\usepackage{algorithmic}

\usepackage{newfloat}
\usepackage{listings}
\DeclareCaptionStyle{ruled}{labelfont=normalfont,labelsep=colon,strut=off} 
\floatstyle{ruled}
\newfloat{listing}{tb}{lst}{}
\floatname{listing}{Listing}
\nocopyright

\title{BPMN to PDDL: Translating Business Workflows for AI Planning}

\author{
    Jasper Nie,
    Christian Muise,
    Victoria Armstrong
}

\affiliations{
    Queen’s University, Canada\\
    20jhn3@queensu.ca,
    christian.muise@queensu.ca,
    victoria.armstrong@queensu.ca
}

\begin{document}

\maketitle

\begin{abstract}
Business Process Model and Notation (BPMN) is a widely used standard for modelling business processes. While automated planning has been proposed as a method for simulating and reasoning about BPMN workflows, most implementations remain incomplete or limited in scope. This project builds upon prior theoretical work to develop a functional pipeline that translates BPMN 2.0 diagrams into PDDL representations suitable for planning. The system supports core BPMN constructs, including tasks, events, sequence flows, and gateways, with initial support for parallel and inclusive gateway behaviour. Using a non-deterministic planner, we demonstrate how to generate and evaluate valid execution traces. Our implementation aims to bridge the gap between theory and practical tooling, providing a foundation for further exploration of  translating business processes into well-defined plans.
\end{abstract}

\section{Introduction}
\label{sec:intro}

Business Process Model and Notation (BPMN) has become the industry standard for modelling complex workflows, offering a visual language for capturing tasks, events, and control flow. Although BPMN provides clarity and facilitates both communication and performing the tasks outlined in the diagram, the diagrams remain primarily descriptive rather than operational. Furthermore, different business processes can be interpreted in a variety of ways due to the nature of how the diagrams are designed to be read, making it valuable to standardize these diagrams. 

A translation from BPMN to automated planning is valuable because it provides analytical capabilities that are not available from BPMN models alone. BPMN diagrams specify how a process is intended to unfold, but they do not check whether the process is executable under all conditions or whether certain tasks can become unreachable, conflicted, or dependent on unintended ordering. A planning-based formulation makes these issues explicit by treating the process as a state-transition system that can be explored and validated computationally. This allows the resulting model to be examined for consistency, completeness, and behavioural soundness, and it also supports the generation of concrete execution sequences that illustrate how the process may evolve in practice.

Automated planning is a core technique in artificial intelligence, and it offers a way to transform static process models into dynamic, analyzable workflows. Previous work~\cite{marrella2019automated,sabatucci2019supporting} has suggested mappings from BPMN to planning domains, but most approaches remain limited in scope, often supporting only basic control-flow elements such as tasks and sequence flows, and frequently stopping at conceptual mappings rather than producing end-to-end, executable planning models.

This project seeks to bridge the gap by developing a functional pipeline that translates BPMN 2.0 diagrams into PDDL representations suitable for classical planning. The system supports core BPMN elements, including tasks, events, sequence flows, and gateways, with initial extensions to message flows as well. By leveraging a leading non-deterministic planner, we obtain planner-generated execution sequences that reflect one possible progression through the translated process. These outputs provide a practical way to examine whether the translation captures the intended behaviour of the BPMN diagram.

The remainder of this paper is structured as follows. Section \ref{sec:prelim} will provide the necessary background on BPMN and automated planning. Section \ref{sec:approach} describes the approach we followed to translate the diagrams into PDDL. Section \ref{sec:illustrative_example} will present an illustrative example of how our system works with a given BPMN diagram. Section \ref{sec:related} will discuss the limitations and directions for future work. Section \ref{sec:concl} concludes with reflections on the contribution and potential applications of this approach.

\begin{figure*}[!t]
    \hfill
    \begin{subfigure}[t]{0.32\textwidth}
        \centering
        \includegraphics[width=\textwidth]{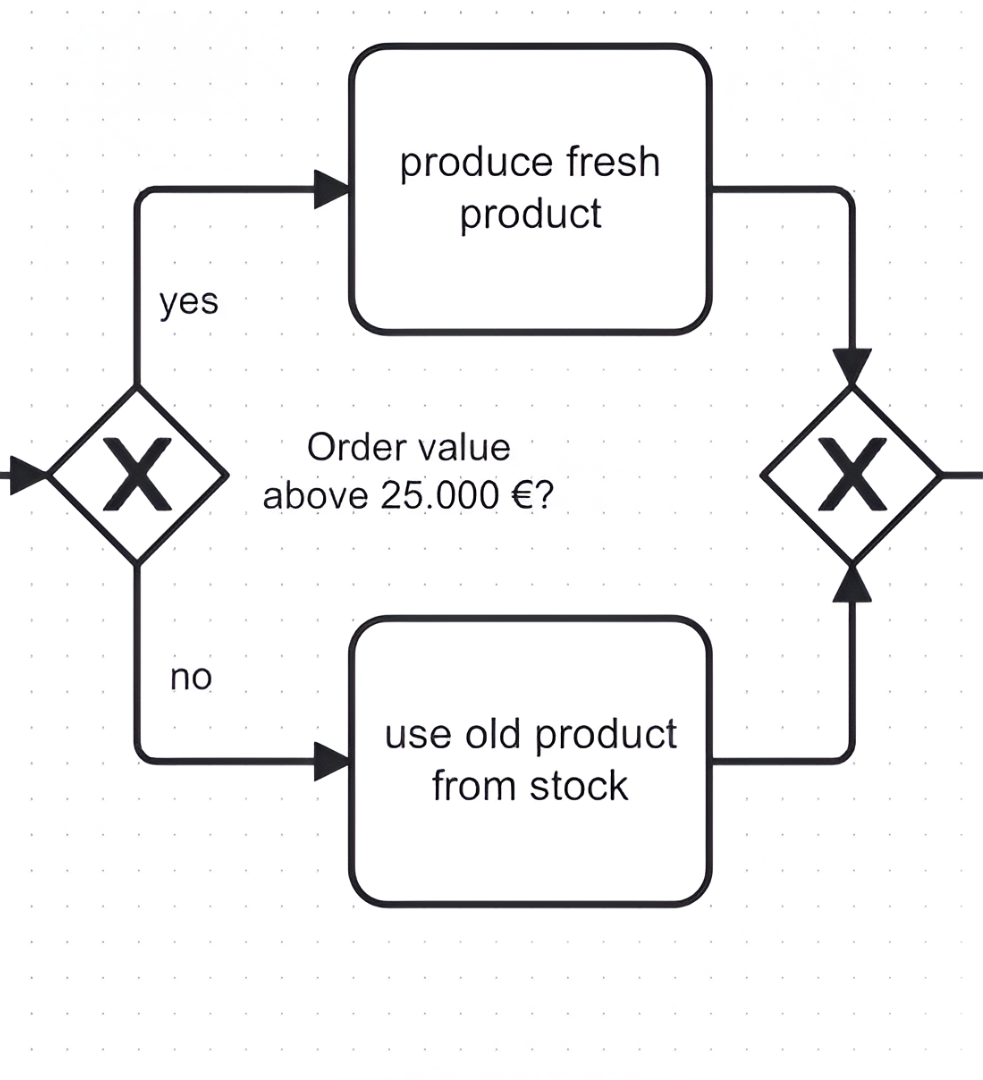}
        \caption{Mutually exclusive effects for exclusive gateways}
        \label{fig:exc}
    \end{subfigure}
    \centering
    \hfill
    \begin{subfigure}[t]{0.32\textwidth}
        \centering
        \includegraphics[width=\textwidth]{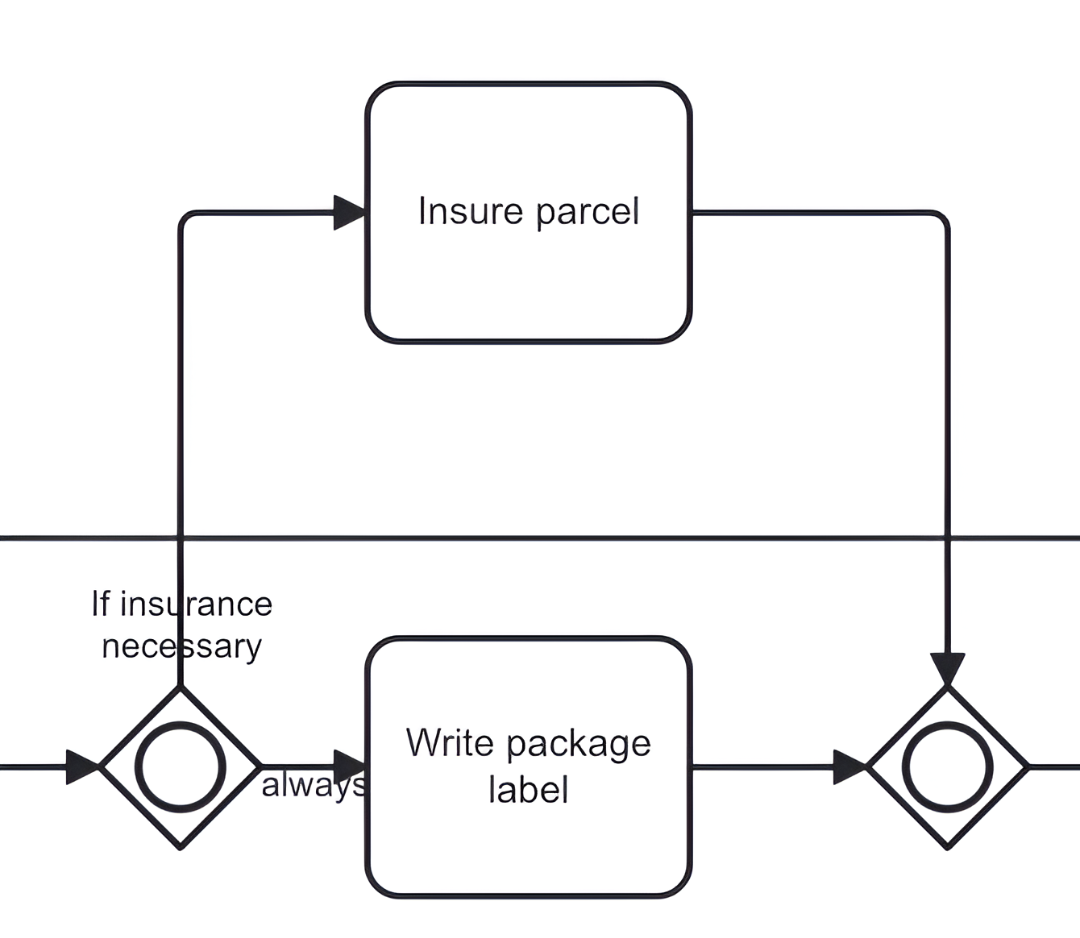}
        \caption{Inclusive gateway with active-branch counters}
        \label{fig:inc}
    \end{subfigure}
    \centering
    \begin{subfigure}[t]{0.32\textwidth}
        \centering
        \includegraphics[width=\textwidth]{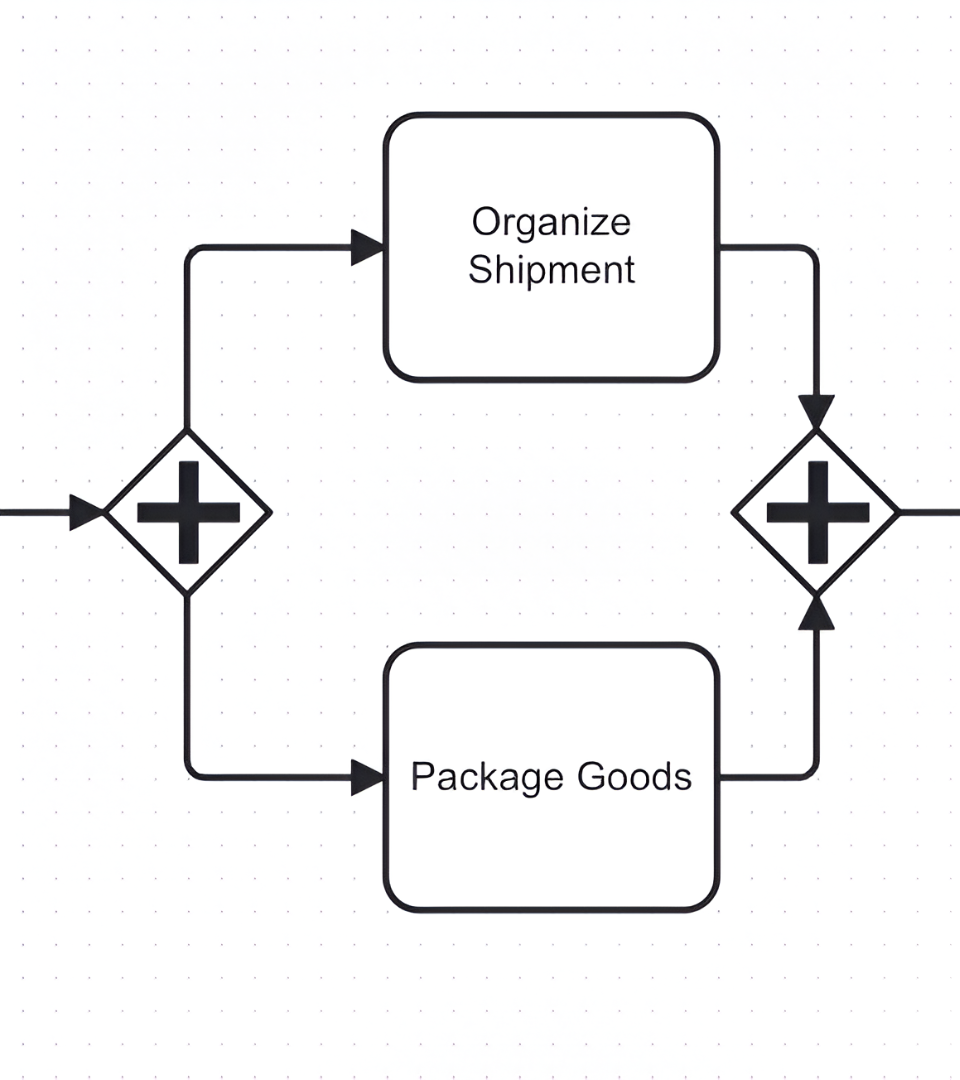}
        \caption{Parallel gateway}
        \label{fig:par}
    \end{subfigure}
    \caption{Examples of parallel, inclusive, and exclusive BPMN gateways}
    \label{fig:gateways}
\end{figure*}
\section{Preliminaries}
\label{sec:prelim}

\subsection{Business Process Model and Notation (BPMN) Diagrams}
Business Process Model and Notation (BPMN) is a standardized graphical language for modelling business processes~\cite{omg2013bpmn}. It provides a common vocabulary for business analysts, system developers, and stakeholders to visualize and reason about workflows. BPMN diagrams are defined using a set of core elements that describe the flow of activities, events, and decisions within a process, together with a formal execution semantics established by the standard~\cite{omg2013bpmn}.

At the most basic level, a BPMN diagram consists of events, activities, and gateways connected by sequence flows~\cite{omg2013bpmn}. \textit{Events} represent occurrences within a process, such as its initiation, the receipt of a message, or its completion. \textit{Activities} correspond to units of work, such as tasks or subprocesses, that are performed during execution. \textit{Gateways} define branching and merging behaviour. Sequence flows indicate the order in which these elements are executed and govern the progression of control between them.

The most critical element for controlling the logic of the workflow is a \textit{gateway}, which directs how sequence flows converge or diverge. BPMN defines several gateway types, each with distinct execution semantics: 

\begin{itemize}
    \item \textbf{Exclusive Gateway (XOR)}: As illustrated in Figure~\ref{fig:exc}, exactly one outgoing branch is taken based on the first condition that evaluates to true according to the gateway’s defined order of evaluation. If no condition is true, a default branch (if specified) is followed.

    \item \textbf{Inclusive Gateway (OR)}: Figure~\ref{fig:inc} shows an inclusive gateway, which allows one or more branches to be taken simultaneously depending on which conditions evaluate to true. It is often used when several activities can proceed in parallel, but not necessarily all.

    \item \textbf{Parallel Gateway (AND)}: As shown in Figure~\ref{fig:par}, a parallel gateway splits the process into multiple branches that execute concurrently. When used as a join, it waits for all incoming branches to complete before continuing.

    \item \textbf{Event-Based Gateway}: Waits for one of several possible events to occur, such as receiving a message or a timer expiring. The gateway then follows the path corresponding to the first event that actually occurs, and all other waiting paths are discarded. For the purposes of the project, the event-based gateway will be treated as an exclusive gateway.
\end{itemize}

A BPMN diagram may also contain \textit{pools}, seen in Figure~\ref{fig:bpmn_diagram}, which represent independent participants or processes within a larger collaboration~\cite{omg2013bpmn}. Each pool may be subdivided into \textit{lanes} (swimlanes), which are used to organize responsibilities or roles within that participant but do not affect the control-flow semantics. Pools execute independently of one another, and interaction occurs only through message flows. In our work, pools are treated as separate processes that are translated individually, with inter-process communication captured through the corresponding message-flow behaviour.

BPMN 2.0 models are stored in XML format, providing a machine-readable representation of process diagrams. Each element in the diagram, such as a task, event, and gateway, are defined as an XML node with attributes specifying its unique ID, name, and type. Connections between elements are represented through \texttt{<bpmn:sequenceFlow>} tags, which determine the starting and ending nodes for each directed edge to capture the flow of control between elements. Additional relationships, such as message flows or data associations, are also encoded in XML, allowing the model to capture both control and communication structures. This structured representation enables automated tools to parse, analyze, and transform BPMN models programmatically, as done in this project's BPMN-to-PDDL translation pipeline.

By combining these elements, BPMN diagrams provide a powerful yet intuitive way to represent both simple and highly complex processes. The standardized semantics of gateways are particularly important, as they ensure that process logic can be interpreted consistently across different modelling and execution platforms.

While BPMN provides a rich notation for business process modelling, it is inherently passive. The diagrams are descriptive rather than executable, capturing the intent and structure of a workflow but not enabling automated reasoning about it.

\subsection{Automated Planning}

Automated planning is a branch of artificial intelligence concerned with finding sequences of actions (or policies) that transform a system from an initial state to a goal state. A planning problem typically consists of:

\begin{itemize}
    \item \textbf{Initial state:} the description of the world before any actions occur.
    \item \textbf{Goal state:} the set of conditions that must hold to consider the problem solved.
    \item \textbf{Actions (operators):} defined by their preconditions (what must be true to apply them) and effects (how they change the world).
\end{itemize}

Planning over deterministic domains treats each action as having exactly one outcome. However, many real-world systems, including workflows with decision points or external events, are non-deterministic. A non-deterministic planner must account for actions that can have multiple possible outcomes and generate plans (or policies) that succeed regardless of which outcome occurs. \cite{ghallab2004automated}

Fully observable non-deterministic (FOND) planning provides a natural fit for modelling BPMN processes, since many BPMN constructs introduce branching behaviour or outcomes that are not fully predictable. In a FOND formulation, the process is represented as a state-transition system where actions may lead to multiple possible successor states. This matches the behaviour of gateways, message-based interactions, and external events in BPMN, all of which can produce alternative execution paths. Because the translation yields a non-deterministic domain, any FOND planner capable of handling branching outcomes can be used to analyze or simulate the resulting workflow.

\subsection{Gap and Opportunity}
While prior research has proposed theoretical mappings from BPMN to automated planning formalisms \cite{gonzalez2013business,marrella2019automated,sabatucci2019supporting}, there remains a clear gap in practical tooling. Most implementations either handle only a subset of BPMN elements-typically tasks and sequence flows, or stop at conceptual demonstrations without producing executable planning domains. As a result, existing approaches do not yet provide a general, end-to-end pipeline that takes arbitrary BPMN 2.0 diagrams as input and produces executable PDDL domains and problems. This limits the ability to evaluate workflows dynamically, simulate execution traces, or analyze alternative process paths in a consistent, reproducible manner.

This project addresses that gap by providing a functional pipeline that translates BPMN diagrams into working PDDL domains. Our system supports core BPMN constructs, including tasks, events, sequence flows, and gateways, with initial support for message flows. By leveraging a non-deterministic planner, the pipeline generates valid execution sequences that preserve the semantics of the original workflow and support practical analysis of the translated BPMN model.
\section{Approach}
\label{sec:approach}

The primary goal of this project is to develop a functional pipeline that converts BPMN 2.0 into PDDL representations suitable for automated planners. The approach is structured to preserve the semantics of BPMN elements while enabling the dynamic execution of workflows through planning techniques. Translating BPMN into the Planning Domain Definition Language (PDDL) transforms these static models into actionable representations that automated planners can process. This translation enables formal analysis, simulation, and plan generation, allowing BPMN workflows to be explored dynamically rather than remaining as fixed diagrams.

In the translation from BPMN to PDDL, tasks are mapped to actions, gateways are represented as conditional or non-deterministic transitions, and message flows or event-based triggers are modelled as potential outcomes. Because of the structure of FOND policies, they can represent the branching structure observed in BPMN diagrams depending on which event or condition occurs, enabling simulation and validation of BPMN models under uncertainty.


\subsection{Mapping BPMN Elements to PDDL}
We map BPMN elements to PDDL as follows:
\begin{itemize}
    \item Tasks: Represented as actions with preconditions corresponding to incoming flows and effects corresponding to outgoing flows. Tasks serve as the main executable units in the generated planning domain.
    \item Events: Start, intermediate, and end events are mapped to predicates indicating their occurrence. Intermediate events may also trigger subsequent actions depending on sequence flows.
    \item Sequence Flows: Represent causal dependencies between tasks and events. In PDDL, they are translated into action effects that enable the next step in the workflow
    \item Message Flows: Represent interactions between elements across different pools. They are converted into additional preconditions and effects to ensure that the sending and receiving elements are properly synchronized. However, only message flows between tasks and events are considered.
\end{itemize}

\subsection{Handling Gateways}
Gateways control the flow of execution and require careful representation in PDDL.
\begin{itemize}
    \item \textbf{Parallel Gateways}: These diverging gateways represent concurrent paths. In PDDL, diverging parallel gateways activate all outgoing branches simultaneously, and converging parallel gateways wait for all incoming branches to complete before proceeding. This ensures that concurrent workflow execution is faithfully simulated.
    \item \textbf{Inclusive Gateways}: Diverging inclusive gateways allow one or more branches to be taken. Converging inclusive gateways wait for all active incoming branches. The PDDL representation uses markers to track which branches are active, allowing the non-deterministic planner to explore all valid combinations of branch execution. To make sure that no branches remain active at the converging gateway, a counter is added to keep track of all of the active branches and there is a check to make sure that the count is zero before continuing.
    \item \textbf{Exclusive Gateways}: Represent choices where exactly one outgoing branch is taken. These are directly translated into mutually exclusive effects in PDDL.
\end{itemize}

\begin{figure*}[!t]
    \centering
    \includegraphics[width=\textwidth]{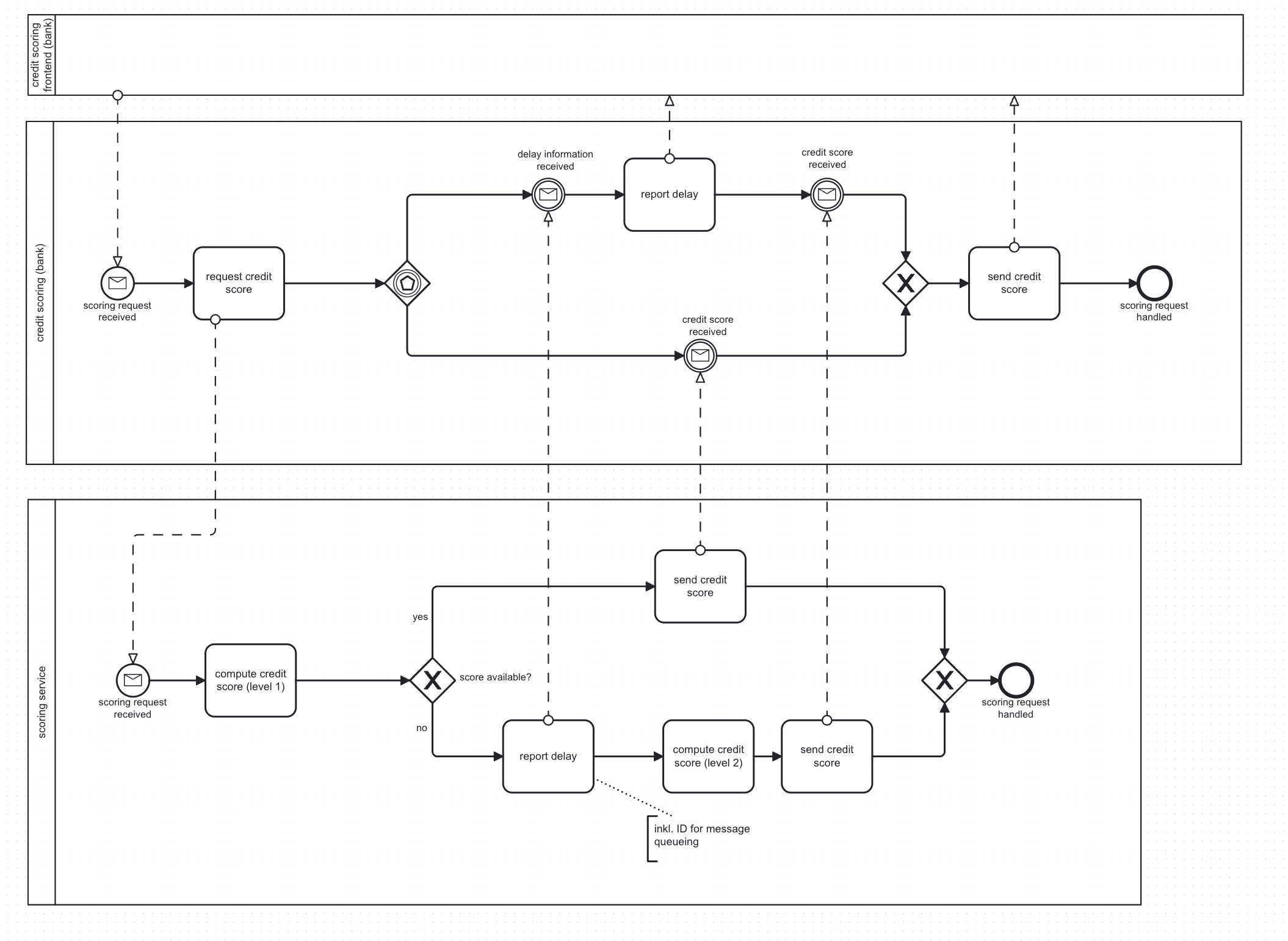}
    \caption{BPMN Diagram}
    \label{fig:bpmn_diagram}
\end{figure*}

\begin{figure*}[!t]
    \centering
    \includegraphics[width=\textwidth]{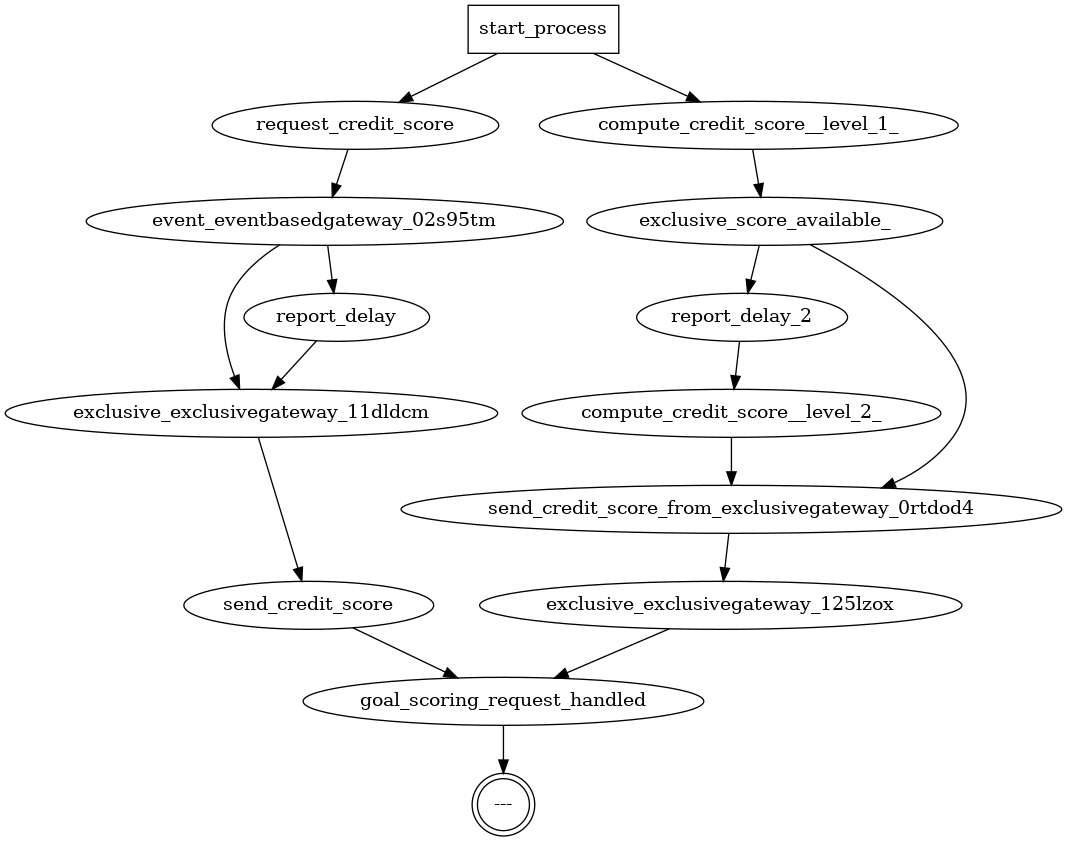}
    \caption{Generated PR2 Policy}
    \label{fig:pddl_policy}
\end{figure*}

We use a FOND planner to demonstrate our faithful translation from BPMN representation to PDDL. Non-determinism is particularly important for inclusive and message flows, where actions have multiple outcomes. 

\subsection{Summary of the Pipeline}
The complete pipeline follows these steps:
\begin{enumerate}
    \item \textbf{Parsing BPMN Diagrams}: BPMN 2.0 XML files are parsed according to the BPMN 2.0.2 specification~\cite{omg2013bpmn}. The implementation uses Python’s \texttt{xml.etree.ElementTree} library~\cite{pythonStdLib} to extract elements such as tasks, events, gateways, sequence flows, message flows, and pools.

    \item \textbf{Constructing Dependencies}: For each BPMN element, incoming and outgoing control-flow relations are identified. This includes the creation of synthetic sequence flows for valid task–event message interactions to ensure they are represented in the control-flow graph.

    \item \textbf{Translating to PDDL}: The extracted BPMN structure is mapped to Planning Domain Definition Language (PDDL) constructs following established planning conventions~\cite{ghallab2004automated}. Predicates, actions, and effects are generated programmatically, with specialized rules for handling gateway semantics, concurrency, and inclusive-branch synchronization.

    \item \textbf{Plan Generation}: Once the domain and problem files are produced, any suitable fully observable non-deterministic (FOND) planner may be applied to compute execution sequences for analysis, validation, or simulation of the translated workflow.
\end{enumerate}

This approach provides a practical, end-to-end solution for translating BPMN diagrams into executable planning representations, bridging the gap between descriptive business process models and automated planning tools.
\section{Illustrative Example}
\label{sec:illustrative_example}

To demonstrate our approach, we present a worked example with an examination of key features. The credit scoring BPMN diagram (Figure~\ref{fig:bpmn_diagram}) models a collaborative process involving three participants: a frontend (bank), a credit scoring service (bank), and an external scoring service. The PDDL translation converts this workflow into a planning domain where process execution becomes a search for valid action sequences.

\subsection{Process Structure}

The process consists of three main pools handling different aspects of credit scoring with timeout mechanisms and two-level score computation. Key elements include tasks (\texttt{request\_credit\_score}, \texttt{send\_credit\_score}, \texttt{report\_delay}), events (start/end events, message catch events), gateways (event-based and exclusive), and cross-participant message flows.

\subsection{Translation Approach}

Each BPMN element becomes a Boolean predicate representing whether that element is currently active in the process state. The PDDL domain begins by declaring the modelling assumptions and introducing types corresponding to the major BPMN classes (tasks, events, and gateways). Figure~\ref{fig:domain-def} shows the header structure of the generated PDDL domain.

\begin{figure}[H]
    \centering
    \begin{minipage}{0.95\columnwidth}
        \begin{lstlisting}[language=PDDL,basicstyle=\ttfamily\footnotesize,xleftmargin=0.5em]
(define (domain credit_scoring)
(:requirements :strips :typing)
(:types task event gateway)
        \end{lstlisting}
    \end{minipage}
    \caption{PDDL domain header and type declaration.}
    \label{fig:domain-def}
\end{figure}

Tasks translate to PDDL actions that capture the basic execution rule of BPMN: a task is executable only when its predecessor element is active, and executing the task transfers control to its successor. The corresponding action encoding is shown in Figure~\ref{fig:request-action}, where the precondition checks that the incoming element is active and the effect activates the next element while deactivating the current one.

\begin{figure}[H]
    \centering
    \begin{minipage}{0.95\columnwidth}
        \begin{lstlisting}[language=PDDL,basicstyle=\ttfamily\footnotesize,xleftmargin=0.5em]
(:action request_credit_score
:precondition (and (StartEvent_1els7eb))
:effect (and (EventBasedGateway_02s95tm)
         (not (StartEvent_1els7eb))))
        \end{lstlisting}
    \end{minipage}
    \caption{Action encoding of \texttt{request\_credit\_score}.}
    \label{fig:request-action}
\end{figure}

Gateway behaviour is encoded using conditional or non-deterministic effects. Event-based gateways wait for one of several possible events, and the first event to occur determines which branch continues. This is modelled using \texttt{oneof}, which specifies that exactly one mutually exclusive outcome will hold after the gateway fires. Exclusive gateways use a similar pattern to represent branching based on routing conditions. Figure~\ref{fig:gateway-action} shows an event-based gateway encoded with non-deterministic successor states.

\begin{figure}[H]
    \centering
    \begin{minipage}{0.95\columnwidth}
        \begin{lstlisting}[language=PDDL,basicstyle=\ttfamily\scriptsize,xleftmargin=0.5em]
(:action event_EventBasedGateway_02s95tm
:precondition (and (EventBasedGateway_02s95tm))
:effect (and (oneof (IntermediateCatchEvent_0ujob24)
             (and (IntermediateCatchEvent_0yg7cuh)
                  (ExclusiveGateway_11dldcm)))
         (not (EventBasedGateway_02s95tm))))
        \end{lstlisting}
    \end{minipage}
    \caption{Event-based gateway with non-deterministic outcomes.}
    \label{fig:gateway-action}
\end{figure}

Cross-participant message flows are converted into synthetic sequence flows during preprocessing. Although message flows do not participate in BPMN control flow, representing them as explicit transitions allows the planner to enforce the ordering constraints that arise when separate pools exchange information, thereby enabling coordinated behaviour across parallel processes.

\subsection{Key Features}

The translation handles: (1) \textbf{Non-determinism} through \texttt{oneof} constructs for exclusive choices, (2) \textbf{State management} via Boolean predicates tracking active elements, (3) \textbf{Cross-process communication} through message flow synthesis, (4) \textbf{Timeout handling} abstracted as non-deterministic event arrival, where race conditions are represented by the planner selecting which event occurs first rather than modelling real temporal behaviour, and (5) \textbf{Process orchestration} enabling coordination across multiple participants.

Different problem instances test various workflow entry conditions: an empty initial state, where the planner autonomously selects which process to initiate; a frontend pre-started scenario, where the customer-facing component of the workflow (e.g., user request handling) is already active; and a scoring service pre-started scenario, where backend evaluation or computation services are initialized before the process begins. These variations allow testing of how the planner handles partial progress or asynchronous components while all instances share the common goal \texttt{(:goal (and (done)))}.

This approach demonstrates how complex business processes with multiple participants, message passing, and conditional logic can be formally modelled as planning problems, providing a foundation for exploring how automated planning techniques may be applied to analyze and interpret BPMN workflows.

Figure~\ref{fig:pddl_policy} shows the policy generated from the translated domain and problem file using the planner PR2~\cite{pr2}. While the example does not aim to optimize execution, the visualized policy confirms that the planner correctly navigates event-based branching, message-triggered transitions, and cross-participant synchronization within the translated model.

\subsection{Other Results}

We evaluated the translation pipeline on a collection of BPMN 2.0 diagrams of varying size and structure, including \texttt{credit\_scoring.bpmn}, \texttt{self\_serve\_restaurant.bpmn}, \texttt{order\_pizza.bpmn}, \texttt{order\_pizza\_2.bpmn}, \texttt{place\_order.bpmn}, \texttt{check\_inventory.bpmn}, \texttt{dispatch\_of\_goods.bpmn}, and \texttt{recourse.bpmn}. These examples cover a range of common BPMN patterns, such as task sequences, exclusive and event-based gateways, parallel behaviour, loops, and cross-participant message flows.

Across all diagrams, translation time was under one second, with generated PDDL domains typically ranging between 50–150 lines, depending on the number of gateways and message interactions. All resulting domains and problem files were accepted by standard FOND planners without modification, indicating that the translation produces syntactically correct and planner-compatible representations. The completely translated PDDL outputs for all diagrams are available in the accompanying GitHub repository. Note most of these examples were chosen to test the tool on specific BPMN patterns and examples. The tool was not tested on full-scale BPMN diagrams mapping out large, complex business processes.

\section{Related Work}
\label{sec:related}

Recent advances in artificial intelligence have significantly impacted business process management, with 2024 marking a pivotal year for AI-driven process automation research. \citeauthor{moreira2024business} (\citeyear{moreira2024business}) conducted a comprehensive systematic literature review on business process automation in small and medium enterprises, highlighting the growing integration of AI technologies with traditional BPM systems. Their work identifies key trends in intelligent automation and establishes a foundation for understanding current BPA implementations.

The intersection of AI and robotic process automation has been extensively explored by Afrin et al.~\cite{afrin2024ai}, who reviewed intelligent automation innovations and their practical applications. Their analysis reveals how AI enhancement transforms traditional RPA from rule-based automation to intelligent, adaptive systems capable of handling complex business scenarios.

From a broader industry perspective, \citeauthor{gurjar2024analytical} (\citeyear{gurjar2024analytical}) provided an analytical review of AI's impact across business sectors, examining applications, emerging trends, and implementation challenges. Their work underscores the transformative potential of AI in revolutionizing business operations while identifying key barriers to adoption.

While these works primarily focus on RPA and general AI applications in business, there remains a gap in systematic approaches to translating formal process models like BPMN directly into planning domain representations. Previous theoretical work~\cite{marrella2019automated,sabatucci2019supporting} has suggested such mappings, but practical implementations that support a broad range of BPMN elements and produce end-to-end PDDL models remain limited.

\section{Conclusion}
\label{sec:concl}

This project presents a functional translation pipeline from BPMN 2.0 to PDDL, enabling automated FOND planners to interpret and simulate business workflows. The system implements an end-to-end parser for BPMN XML files that generates PDDL domains and problem instances representing the behaviour of core BPMN constructs such as tasks, events, and gateways, with initial extensions for parallel, inclusive, and message flow behaviour. The translation was validated through a working example described in Section 4, where multiple problem instances were generated to demonstrate correct gateway synchronization, message flow handling, and execution-trace generation using the PR2 planner. This work establishes a practical and extensible foundation that connects business process modelling with automated planning, bridging the gap between descriptive diagrams and executable reasoning systems.

\subsection{The Message Flow Problem}
Currently, message flows between a task and an event are treated as sequence flows, while message flows between two tasks are ignored as simple communication links. As a result, tasks with outgoing message flows behave similarly to parallel gateways, since both the normal sequence flow and the message flow should activate subsequent elements. However, unlike parallel gateways, message flows lack a converging gateway to ensure that all branches are explored. This causes the planner to execute tasks sequentially rather than exploring multiple paths. Our current solution treats these message flows as exclusive gateways, leveraging the non-deterministic nature of the PR2 planner to force branching behaviour in the generated policy. While this approach allows message flows to be explored correctly, it does not fully preserve their intended semantics, highlighting a key limitation to be addressed in future work.

\subsection{Limitations and Unimplemented Elements}
Although the current system supports the core behavioural components of BPMN 2.0, such as tasks, events, sequence flows, and gateway types (inclusive, exclusive, and parallel), some elements remain unimplemented. Complex constructs like subprocesses, boundary and timer events, data objects, signal flows, and compensation mechanisms are not yet translated into PDDL. These components often introduce additional contextual dependencies or asynchronous behaviours that require specialized handling to preserve BPMN semantics during planning translation.

Furthermore, the current parser focuses primarily on control-flow semantics rather than data flow or resource allocation. As a result, aspects such as message correlation, actor-specific ``swimlanes'', and transactional behaviour between participants are simplified or omitted. Extending the system to incorporate these missing constructs will be essential to achieving a complete BPMN-to-PDDL translation framework, enabling planners to simulate not only task execution but also the rich interactivity and concurrency that define real-world business processes.

\bigskip

\bibliography{aaai25}

@inproceedings{pr2,
  author       = {Christian Muise and
                  Sheila A. McIlraith and
                  J. Christopher Beck},
  editor       = {Michael J. Wooldridge and
                  Jennifer G. Dy and
                  Sriraam Natarajan},
  title        = {{PRP} Rebooted: Advancing the State of the Art in {FOND} Planning},
  booktitle    = {Thirty-Eighth {AAAI} Conference on Artificial Intelligence, {AAAI}
                  2024, Thirty-Sixth Conference on Innovative Applications of Artificial
                  Intelligence, {IAAI} 2024, Fourteenth Symposium on Educational Advances
                  in Artificial Intelligence, {EAAI} 2014, February 20-27, 2024, Vancouver,
                  Canada},
  pages        = {20212--20221},
  publisher    = {{AAAI} Press},
  year         = {2024},
  url          = {https://doi.org/10.1609/aaai.v38i18.30001},
  doi          = {10.1609/AAAI.V38I18.30001},
  bibsource    = {dblp computer science bibliography, https://dblp.org}
}

@techreport{omg2013bpmn,
  author       = {{Object Management Group}},
  title        = {Business Process Model and Notation (BPMN) Version 2.0.2},
  institution  = {Object Management Group (OMG)},
  year         = {2013},
  url          = {https://www.omg.org/spec/BPMN/2.0.2/},
  note         = {Formal specification defining the standard semantics for BPMN elements and gateways.}
}

@book{ghallab2004automated,
  title     = {Automated Planning: Theory and Practice},
  author    = {Malik Ghallab and Dana Nau and Paolo Traverso},
  year      = {2004},
  publisher = {Morgan Kaufmann},
  isbn      = {978-1-55860-856-6}
}

@article{marrella2019automated,
  title={Automated planning for business process management},
  author={Marrella, Andrea},
  journal={Journal on Data Semantics},
  volume={8},
  number={2},
  pages={79--98},
  year={2019},
  publisher={Springer}
}

@article{sabatucci2019supporting,
  title={Supporting dynamic workflows with automatic extraction of goals from BPMN},
  author={Sabatucci, Luca and Cossentino, Massimo},
  journal={ACM Transactions on Autonomous and Adaptive Systems},
  volume={14},
  number={2},
  pages={1--31},
  year={2019},
  publisher={ACM}
}

@article{moreira2024business,
  title={Business process automation in SMEs: a systematic literature review},
  author={Moreira, S{\'e}rgio and Mamede, Henrique S and Santos, Andr{\'e}},
  journal={IEEE Access},
  volume={12},
  pages={70691--70707},
  year={2024},
  publisher={IEEE},
  doi={10.1109/ACCESS.2024.3404621}
}

@article{afrin2024ai,
  title={AI-enhanced robotic process automation: A review of intelligent automation innovations},
  author={Afrin, Shahana and Roksana, Sohana and Akram, Rashida},
  journal={IEEE Access},
  volume={12},
  pages={186534--186552},
  year={2024},
  publisher={IEEE},
  doi={10.1109/ACCESS.2024.3518237}
}

@article{gurjar2024analytical,
  title={An analytical review on the impact of artificial intelligence on the business industry: Applications, trends, and challenges},
  author={Gurjar, Kapil and Jangra, Ankit and Baber, Hashem and Islam, Md Mominul and others},
  journal={IEEE Engineering Management Review},
  volume={52},
  number={2},
  pages={45--58},
  year={2024},
  publisher={IEEE},
  doi={10.1109/EMR.2024.3365982}
}

@article{gonzalez2013business,
  title={From business process models to hierarchical task network planning domains},
  author={Gonz{\'a}lez-Ferrer, Antonio and Fern{\'a}ndez-Olivares, Juan and Castillo, Luis},
  journal={The Knowledge Engineering Review},
  volume={28},
  number={2},
  pages={175--194},
  year={2013},
  publisher={Cambridge University Press},
  doi={10.1017/S0269888912000409}
}

@misc{pythonStdLib,
  title        = {The Python Standard Library},
  author       = {{Python Software Foundation}},
  year         = {2024},
  howpublished = {\url{https://docs.python.org/3/library/xml.etree.elementtree.html}},
  note         = {Accessed: 2025-02-18}
}

\end{document}